\pdfoutput=1
\documentclass[11pt]{article}

\usepackage{acl}

\usepackage{times}
\usepackage{latexsym}

\usepackage[T1]{fontenc}
\usepackage[utf8]{inputenc}

\usepackage{microtype}

\usepackage{listings}
\usepackage{xcolor}
\usepackage{tabularx}
\usepackage{graphicx}
\usepackage{amsmath}
\usepackage{multirow}
\usepackage{qtree}

\title{Tree Transformers are an Ineffective Model of Syntactic Constituency}

\author{Michael Ginn \\
  University of Colorado\\
  \texttt{michael.ginn@colorado.edu}}

\begin{document}
\maketitle
\begin{abstract}
Linguists have long held that a key aspect of natural language syntax is the recursive organization of language units into constituent structures, and research has suggested that current state-of-the-art language models lack an inherent bias towards this feature. A number of alternative models have been proposed to provide inductive biases towards constituency, including the Tree Transformer \citep{Wang2019TreeTI}, which utilizes a modified attention mechanism to organize tokens into constituents. 

We investigate Tree Transformers to study whether they utilize meaningful and/or useful constituent structures. We pretrain a large Tree Transformer on language modeling in order to investigate the learned constituent tree representations of sentences, finding little evidence for meaningful structures. Next, we evaluate Tree Transformers with similar transformer models on error detection tasks requiring constituent structure. We find that while the Tree Transformer models may slightly outperform at these tasks, there is little evidence to suggest a meaningful improvement. In general, we conclude that there is little evidence to support Tree Transformer as an effective model of syntactic constituency. 
\end{abstract}

\section{Introduction}
Linguists have long believed that humans utilize hierarchical and recursive structure in natural language, allowing for highly productive generalizations and rapid acquisition of complex patterns \citep{chomsky1980, LegateYang+2002+151+162, crain1987}. \textbf{Hierarchical structure} organizes words into \textit{constituents}, phrases that act as single units and can be composed into larger constituents, allowing for patterns such as English possessive phrases (\textit{the man over there's dog}). \textbf{Recursive structure} occurs when rules can be repeatedly applied, such as English relative clause embedding (\textit{the food that the cat that the dog chased ate}). According to the \textbf{poverty of the stimulus} argument, humans have an inherent bias towards productive hierarchical and recursive patterns, allowing them to learn complex natural language rules from limited data \citep{chomsky1980}. 

\begin{figure}
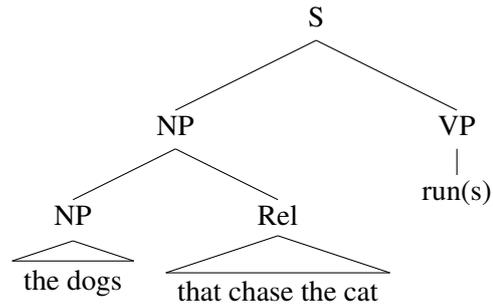

    \centering
    % \includegraphics{Tree.eps}
    % \Tree [.S [.NP \qroof{the dogs}.NP ][.VP run(s)]]
    \Tree [.S  [.NP \qroof{the dogs}.NP \qroof{that chase the cat}.Rel ] [.VP run(s) ] ]
    \caption{Tree for a sentence that requires a hierarchical generalization to label correctly}
    \label{fig:tree}
\end{figure}

Meanwhile, modern NLP approaches generally utilize Transformer-based language models (TLMs) that operate on unstructured text, learning continuous latent representations of words in context in order to perform tasks. In many cases, these vector representations can approximate the hierarchical relationships proposed to exist in natural language \citep{jawahar-etal-2019-bert, peters_deep_2018}, however, these representations may not generalize well beyond the training data \citep{anil2022exploring, hu2020systematic}. Furthermore, language models are able to learn patterns that do not occur in natural language \citep{MORO202382, kallini_mission_2024}.

Recent work has found that various approaches that utilize \textbf{explicit hierarchical representations} can improve performance and generalization across tasks such as learning syntactic transformations \citep{mccoy_does_2020}, semantic role labeling \citep{li_unified_2018}, and coreference resolution \citep{jiang_incorporating_2021, kong_incorporating_2019}. These approaches generally depend on canonical tree parses for optimal performance, and noisy tree representations can impair effectiveness. Thus, it remains an open question whether models that automatically infer tree structures can also model language better than standard TLMs. 

\subsection{Hierarchical Language Models}
In this work, we explore architectures that infer \textbf{latent hierarchical representations}, without requiring expert parses, as a model of syntactic structure. Specifically, we train models using the Tree Transformer architecture \citep{Wang2019TreeTI}, which modifies the attention mechanism in standard Transformers with an additional \textit{constituent attention} mechanism. Constituent attention learns which tokens should form constituents, and constrains standard attention so that tokens can only attend to other tokens in the same constituent. This process is repeated at each level, inducing a tree-like representation where constituents grow larger and larger. 

We investigate the effectiveness of Tree Transformers with two sets of experiments. In \autoref{sec:pretraining}, we pretrain large language models using the Tree Transformer architecture on an unlabeled English dataset. We investigate the learned representations in a systematic way, analyzing how the model treats syntactic units such as determiners, adjectives, and relative clauses. We compare this model with typical linguistic beliefs about syntactic structure, and find that while the model follows some of these beliefs, it often learns syntactic parses that are highly unexpected and seem unlikely to be linguistically meaningful.

In \autoref{sec:agreement}, we evaluate these models on the task of predicting syntactic violations: given a sentence, predict whether it is syntactically well-formed or contains some violation. We design experimental settings where the model must learn to make a hierarchical generalization to produce the correct answer, with sentences such as \autoref{fig:tree}. We compare Tree Transformer models with equivalent Transformers, finding that the pretrained Tree Transformers make small improvements at making correct generalizations, but far from the benefit that we would expect from a model using tree-structured data. We also find that, in contrast to prior research, pretraining for non-hierarchical models does not necessarily impart the necessary hierarchical bias. Overall, we fail to find evidence that Tree Transformer is a more effective model of language than an equivalent transformer. Our code is available on GitHub\footnote{\url{https://github.com/michaelpginn/latent-trees}}.

\section{Background}
\subsection{Large Language Models}
Large Language Models (LLM) are the state-of-the-art across many NLP tasks, with notable examples including BERT \citep{devlin2019bert} and GPT \citep{radford2018improving, radford2019language, brown2020language}. LLMs are pretrained on raw text corpora to predict missing words, requiring the model to learn complex representations of words and phrases that allow for contextual prediction. LLMs may then be \textit{finetuned} on a particular task, often outperforming models trained directly on the task \citep{devlin2019bert}.

Modern LLMs use the Transformer architecture \citep{vaswani2017attention}, which learns large, continuous vector representations of tokens and utilizes an \textit{attention mechanism} that models the interaction between tokens. These representations have been shown to learn linguistic features including syntactic and semantic information \citep{jawahar-etal-2019-bert, peters_deep_2018}, however, interepretation remains difficult. We train LLMs using a modified architecture, Tree Transformer, which captures discrete relationships between words and produces hierarchical representations of sentences. 

\subsection{English Subject-Verb Agreement}
\label{sec:heuristics}
In \autoref{sec:agreement}, we evaluate models on the binary classification task of detecting agreement violations in English. Though previous work \cite{mccoy_does_2020, petty_Transformers_2021, wilson-frank-2023-inductive} has used sentence-to-sentence transformation tasks (such as question formation), models that must generate entire sentences are prone to errors unrelated to the generalization being studied, making evaluation more difficult. 

Humans display a hierarchy-sensitive rule for subject-verb agreement. For example, given the sentences \textit{"the dogs run"} and \textit{"the dog that chases the cat runs"}, there are two possible agreement rules:
\begin{enumerate}
    \item The verb should agree with the nearest noun on the left side (linear heuristic)
    \item The verb should agree with the subject of the verbal phrase (hierarchical heuristic)
\end{enumerate}

Under the linear heuristic, we might predict the incorrect sentence \textit{"the dogs that chase the cat runs"}, while only the hierarchical heuristic produces the correct sentence \textit{"the dogs that chase the cat run"}. We refer to the application of such a hierarchical rule as \textbf{hierarchical generalization}.

\autoref{fig:tree} shows a partial tree for this example, demonstrating one possible hiearchical structure that would allow for learning the correct generalization.

The agreement rule can be applied to arbitrarily recursive structures. For example, English allows recursive embedding of relative clauses, as in \textit{"The men that work on the bridge that the city owns are chatting"}. Though humans rarely apply these recursive rules more than three or four times \citep{karlsson2007constraints}, they can theoretically ensure agreement is valid across arbitrarily deep structures. We refer to this ability as \textbf{recursive generalization}. 

\subsection{Related Work}
\subsubsection{Models with hierarchical biases}
\label{sec:other_models}
A number of model architectures have been proposed to better handle hierarchical and recursive patterns. This is not an exhaustive list of models of the sort, but covers many of the key approaches.

\citet{Wang2019TreeTI} proposes Tree Transformers, a drop-in replacement for the standard attention mechanism, where an additional \textit{constituent attention} mechanism limits tokens to only attend to other tokens in the same constituent at each self-attention layer. We study this architecture in this research and describe the details in \autoref{sec:tree-transformer}.

Ordered-Neurons LSTM \citep{shen_ordered_2019} learns an ordering that forces neurons to represent information for varying lengths of time, resulting in low-ranking neurons that learn local information and form constituents.

Recursive Neural Networks (RvNNs, \citealt{pollack1990recursive, socher2010learning}) repeatedly apply a neural operation in order to compose items within a sequence. However, these models require a predetermined tree order for composing items, and cannot induce this representation. Gumbel-Tree LSTMs \citep{Choi_Yoo_Lee_2018} induce latent tree structures using straight-through estimation in order to train the discrete task of selecting tree structures. CRvNN \citep{chowdhury_modeling_2021} relaxes the decision function to predict composition probabilities and thus allow for end-to-end differentiable training. 

Finally, reinforcement learning has been used to learn to create tree structures using performance on a downstream task as a reward signal \citep{yogatama2016learning, romer2020reinforcement}. 

We focus on the Tree Transformer of \citet{Wang2019TreeTI}, as it provides a direct comparison to existing language models and can be easily adopted in state-of-the-art architectures.

\subsubsection{Evaluating model generalization}
This work is inspired by \citet{mccoy_does_2020}, which studies the ability of Transformers to make hierarchical generalizations. They find that standard recurrent models fail to make the correct generalization from ambiguous data, while graph neural models that use explicit syntax trees as input, such as Tree-GRU \citep{chen-etal-2017-improved}, learn the correct generalizations. \citet{petty_Transformers_2021, yedetore_how_2023, wilson-frank-2023-inductive} have similar findings for Transformer-based models, which tend to select either linear heuristics or rules based on counting. \citet{deletang_neural_2023} generalizes this result by studying the ability of RNNs and Transformers to learn tasks at various levels of the Chomsky hierarchy, finding that these models are unable to reliable learn any non-regular patterns. However, while these works find limitations in the ability of common neural models to make hierarchical generalizations, none of them study alternate architectures that generalize more effectively without using explicit structural information (expert-labeled parse trees).

Meanwhile, \citet{mueller_coloring_2022, mueller_how_2023, warstadt_can_2020} study pretrained language models, finding that pretraining can impart a hierarchical bias. However, this ability may depend on the pretraining corpus, with \citet{mueller_how_2023} finding that models learn to make hierarchical generalizations more readily when trained on a curriculum that initially uses simpler language (child-directed speech). \citet{yao_self-attention_2021} finds that self-attention models can learn recursive formal languages, so long as recursion has a finite bound. 

\subsubsection{Grammar Induction}
This work is closely related with the topic of \textit{grammar induction}, in which grammatical parses are extracted from data in an unsupervised manner. Grammar induction has a rich history, with approaches including iterative trial and error \citep{duda_pattern_2001}, genetic algorithms \citep{whigham_grammatical_1996}, and probabilistic context-free grammars \citep{kim_compound_2019}. Grammar induction in the context of neural networks has been explored through networks that use hierarchical latent representations such as ON-LSTM \citep{shen_ordered_2019}, latent neural grammars \citep{kim_sequence--sequence_2021}, and sequence-to-dependency networks for translation \citep{wu_sequence--dependency_2017}. 

\section{Tree Transformer}
\label{sec:tree-transformer}
The Tree Transformer model \citep{Wang2019TreeTI} uses a modification to the attention mechanism to induce tree hierarchy from input tokens. Tree Transformer induces constituents at each level, and restricts the attention mechanism so only tokens in the same constituent may attend to each other. Constituents strictly grow in size with each additional layer, until all words are merged into a single constituent. 

This mechanism depends on a \textit{constituent prior}, $C$, an $N \times N$ matrix where $C_{i,j}$ denotes the probability that word $i$ is in the same constituent as word $j$. The standard attention probability matrix is constrained by the constituent probabilities at each word, as in \autoref{eq:attention}, where $Q$ is the query vector for each token, $K$ is the key vector, and $d$ is the hidden dimension.

\begin{equation}
E = C \odot \textnormal{softmax}(\frac{QK^T}{d})
\label{eq:attention}
\end{equation}
The same constituency matrix is used across attention heads. To calculate the constituency matrix, we calculate a score $s_i$ for linking each token to the left or right token, using an additional key and query vector, with learned weights (\autoref{eq:constituent-prior}).
\begin{equation}
s_{i, i+1} = \frac{q_i \cdot k_{i+1}}{d}
\label{eq:constituent-prior}
\end{equation}

Next, we take the softmax over the two possible links to encourage only one merge and convert scores to link probabilities $p_{i, i+1}$ and $p_{i, i-1}$ (\autoref{eq:softmax}).
\begin{equation}
p_{i, i+1}, p_{i, i-1} = \textnormal{softmax}(s_{i, i+1}, s_{i, i-1})
\label{eq:softmax}
\end{equation}

We take the geometric mean between the leftward and rightward link probabilities, which gives a probability $a_i$ that word $i$ and $i+1$ are in the same constituent (\autoref{eq:mean}).

\begin{equation}
    \hat{a_i} = \sqrt{p_{i, i+1} \times p_{i, i-1}}
    \label{eq:mean}
\end{equation}

We enforce that the probability at any layer $a_k^{l}$ must be strictly larger than the probability at the previous attention layer $a_k^{l-1}$, so words merge together into larger and larger constituents (\autoref{eq:monotonic_increase}).

\begin{equation}
    a^l_k = a^{l-1}_k + (1-a^{l-1}_k)\hat{a}^l_k
    \label{eq:monotonic_increase}
\end{equation}

Finally, we calculate the constituent prior $C_{i,j}$ between all pairs of tokens at $i, j$ using the $a$ probabilities (\autoref{eq:constituent_prior_calc}).

\begin{equation}
    C_{i,j} = \prod^{j-1}_{k=i} a_k
    \label{eq:constituent_prior_calc}
\end{equation}

We found that using the code from \citet{Wang2019TreeTI} directly, even without the custom attention mechanism, was unable to match the performance of the HuggingFace models, indicating that there were differences in optimization techniques. Thus, we reimplemented this mechanism in the HuggingFace framework, allowing it to be utilized as a drop-in addition for any Transformer architecture. For this research, we used the Tree Transformer mechanism with the standard BERT architecture.

The constituent attention mechanism can also be thought of as enforcing lower entropy of attention weights at lower levels, as tokens are encouraged to attend to a few other tokens strongly, and greater entropy at higher levels, where constituents are very large. Essentially, this creates a gradient of the density of attention weights. The constituent attention mechanism does add additional parameters, however, it is quadratic with respect to sequence length and essentially equivalent to adding an attention head.

It is important to note that this architecture does not enforce discrete syntax trees, but uses constituent probabilities to approximate a tree-like structure. Future work could explore modifications that utilize true symbolic trees, while maintaining differentiability through an approach such as straight-through gradient estimation \citep{bengiostraight}.

\subsection{Tree Parsing}
\label{sec:parsing}
We utilize trained Tree Transformer models to create parse trees in the manner specified in \citet{Wang2019TreeTI}. We start at the highest layer in the model and select the token with the lowest constituency probability ($a^l_k$). If this probability is less than some threshold, we split the sentence at that word, and move to the token with the next lowest probability. If there are no other tokens in the current layer to split, we move to the next layer and repeat. 

\section{Exploring a Tree LLM}
\label{sec:pretraining}
In order to study the sorts of representations learned by a Tree Transformer model, we pretrain an LLM using the architecture on a large corpus. In particular, we replicate the training of BERT described in \citet{devlin2019bert}. 

\subsection{Pretraining}
Our model uses the same architecture as \textsc{BERT-Base}, with 12 encoder layers, a hidden size of 768, 12 attention heads, and 124M total parameters. Like BERT, our model is trained on a large English corpus consisting of the BooksCorpus \citep{zhu2015aligning} and English Wikipedia, with a total of around 3.3B words. The data is tokenized using WordPiece \citep{wu2016google}, which divides words into subword units and allows the model to handle out-of-vocabulary words. Due to resource constraints, we are unable to pretrain on this entire corpus, so we evenly sample training batches to use for pretraining with a total of 51M tokens; this is far more data than the original training corpus of 1.1M tokens used in \citet{Wang2019TreeTI}. The data is packed into sequences of 512 tokens, adding padding as needed.

Our LLM is trained using \textit{masked language modeling}, where tokens are randomly replaced with either mask tokens or random tokens, and the model must recover the original text. We pretrain using the hyperparameters listed in \autoref{tab:pre_hyperparameters}. Training took around 2 days using an NVIDIA V100 GPU.

\begin{table}[htb]
    \def\arraystretch{1.5}
    \centering
    \begin{tabularx}{\columnwidth}{|X | c|}
    \hline
    Parameter & Value \\
    \hline
    Optimizer & AdamW  \\
    $\beta_1$ & 0.9 \\
    $\beta_2$ & 0.999 \\
    $\epsilon$ & $1\text{E}{-8}$ \\
    Weight deecay & 0.01 \\
    Learning rate & $2\text{E}{-5}$ \\
    Batch size & 192 \\
    Max Epochs & 40 \\
    \hline
    \end{tabularx}
    \caption{Pretraining hyperparameters}
    \label{tab:pre_hyperparameters}
\end{table}

Our pretrained model, which we refer to as TreeBERT, is available on HuggingFace\footnote{\url{https://huggingface.co/michaelginn/treebert-pretrained-100k}}.

\subsection{Methodology}
\begin{figure*}[h]
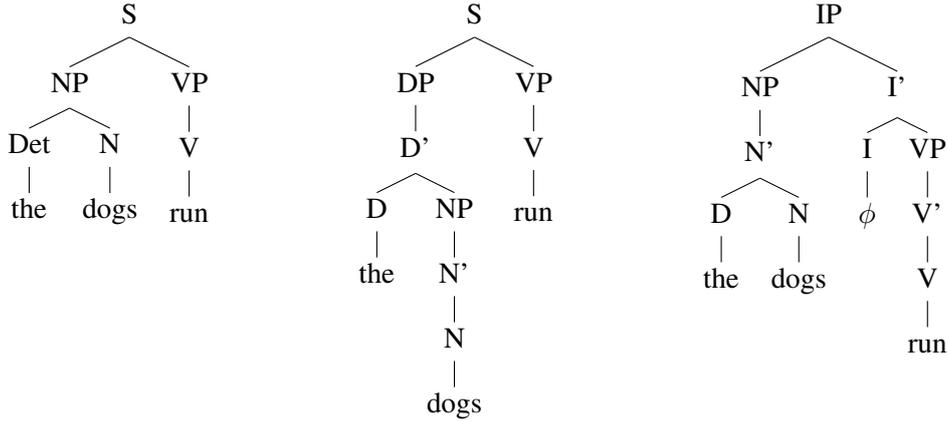

    \centering
        \Tree [.S  [.NP [.Det the ] [.N dogs ] ] [.VP [.V run ] ] ]
        \hspace{1cm}
        \Tree [.S  [.DP [.D' [.D the ] [.NP [.N' [.N dogs ] ] ] ] ] [.VP [.V run ] ] ]
        \hspace{1cm}
        \Tree [.IP  [.NP [.N' [.D the ] [.N  dogs ] ] ] [.I' [.I $\phi$ ]  [.VP [.V' [.V run ] ] ] ] ]
        \hspace{1cm}
    \caption{Possible tree parses for \textit{"the dogs run"} under various syntactic theories}
    \label{fig:dogtree}
\end{figure*}

We produce parses for each sentence using the trained model as described in \autoref{sec:parsing}. We are interested in whether the tree representations correspond with common linguistic insights about syntactic structure.

Since possible parses could take many structures, we explore properties based on \textit{precedence}, studying which words tend to be merged first. In particular, we examine the behavior of the model for determiners, adjectives, and relative clauses. Though syntactic theories differ in the exact predicted structure, there are common hypotheses made about these syntactic units. For instance, \autoref{fig:dogtree} shows the structural interpretation of the sentence \textit{"the dogs run"} under several syntactic theories. Though the structures differ, the majority of theories share the feature that \textit{the} is merged with \textit{dogs} before \textit{dogs} is merged with \textit{run}. We explore whether these common predictions are held in the Tree Transformer latent representations. 

Furthermore, as the induced trees are defined by a series of probabilities, there are many cases where the model assigns high probabilities to multiple possible structures, and there was significant variance in the exact tree structures selected. Thus, it was insufficient to examine individual trees, which might not be indicative of the overall trend; rather, we study the average tendencies of the model by analyzing many generated tree structures for different sentences.

\subsection{Results}
We observe initially that the breakpoint probabilities ($a_k$) tend to be very close to 50\% at all tokens in the lowest layer, 75\% in the second layer, 87.5\% in the third, and so on, halving the distance each time. This indicates that the only differences in breakpoint probabilities are small, and the model does not learn strong absolute preferences in hierarchical structure. By the fourth layer and above, the model is essentially identical to a standard transformer. In \citet{Wang2019TreeTI}, the lowest few layers contained slightly more varied break probabilities–we speculate that as we trained on a much larger corpus (with packed input sequences), these probabilities were regularized. However, over many trials we do observe clear preferences resulting from small differences in lower level constituent probabilities, which we investigate here.

\paragraph{Determiners}
We study simple sentences of the form "Det N V (NP)". We generate sentences using a small vocabulary of 15 nouns, 5 intransitive verbs, and 6 transitive verbs. We create all possible sentences including singular and plural nouns and transitive and intransitive verbs, and generate 5,550 tree parses. We analyze whether the determiner and subject are merged first, giving the standard structure "[[Det N] VP]", or whether the subject and verb are merged first, giving an unexpected structure "[Det [N VP]]". Our results are given in \autoref{tab:determiners}.

\begin{table}[h]
    \def\arraystretch{1.5}
    \centering
    \begin{tabular}{|c|c c|}
    \hline
    Sentence type & [Det N] & [N VP] \\
    \hline
         Sing. subject, intrans. & 68 & 7  \\
         Plur. subject, intrans. & 72 & 3 \\
         Sing. subject, trans. & 2376 & 324 \\
         Plur. subject, trans. & 2642 & 58 \\
    \hline
    \end{tabular}
    \caption{Counts for sentence parses that merge the determiner and subject first or subject and verb first}
    \label{tab:determiners}
\end{table} 

We see that the model shows a strong preference toward merging determiners and nouns. With completely random binary merges, the null hypothesis would be that either option is equally likely and the "[Det N]" merge would happen 50\% of the time. It is clear (with $p<1\mathrm{e}{-30}$) that the tree parses are different from random binary merges. The result aligns with linguistic intuitions about the division between noun phrases and verb phrases. From a language modeling perspective, one explanation is that in order to predict whether there is a determiner, the model must consider the type of noun first, while forming a constituent with the noun and verb would make it more difficult to recover this information.

\paragraph{Adjectives}
We follow the procedure of the previous section, adding a vocabulary of 8 adjectives to create sentences of the form "Det Adj N VP (NP)". For the transitive sentences, we randomly sample 5,400 sentences, as the number of possible sentences is very large. We investigate whether the model tends to form a constituent with the determiner and adjective or adjective and noun, and observe the results in \autoref{tab:adjectives}.

\begin{table}[h]
    \def\arraystretch{1.5}
    \centering
    \begin{tabular}{|c|c c|}
    \hline
    Sentence type & [Det Adj] & [Adj N] \\
    \hline
         Sing. subject, intrans. & 476 & 124  \\
         Plur. subject, intrans. & 500 & 100 \\
         Sing. subject, trans. & 4131 & 1269  \\
         Plur. subject, trans. & 4683  & 717  \\
    \hline
    \end{tabular}
    \caption{Counts for sentence parses that merge the determiner and adjective first or adjective and noun first}
    \label{tab:adjectives}
\end{table} 

\begin{figure}
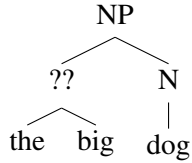

    \centering
    % \includegraphics{Tree.eps}
    % \Tree [.S [.NP \qroof{the dogs}.NP ][.VP run(s)]]
    \Tree [.NP  [.?? the big ] [.N dog ] ]
    \caption{Common structure predicted by Tree Transformer for noun phrases with adjectives}
    \label{fig:adjective_tree}
\end{figure}

We observe a strong preference toward merging determiner and adjectives before adjectives and nouns. This is an unexpected result---standard generative syntax tends to suggest that adjectives and nouns form a constituent, allowing for recursive adjective attachment in phrases such as "big happy red dog". Meanwhile, the Tree Transformer model generally prefers structures such as \autoref{fig:adjective_tree}.

As Tree Transformer utilizes learned constituency to regulate attention, it is unclear how attending between a determiner and adjective would be more beneficial than attending between an adjective and noun (which have a clear semantic interaction). Furthermore, this tendency follows the ordering of the prior one–it seems possible that Tree Transformer simply prefers merging the first two words of a sentence, based on some statistical pattern that is not supported by linguistic theory. 

% It is unclear exactly why the model might prefer this structure, as there is no obvious relationship between adjectives and determiners. Prior research has shown that higher layers of transformers tend to represent semantic information, while lower layers represent syntactic relationships \citep{jawahar-etal-2019-bert}. Considering this, a possible explanation is that adjectives are more critical for semantic structures than syntactic ones, so they are merged with nouns at higher layers. If this is the case, it is indicative of a limitation of the Tree Transformer model, where layers in the model must simultaneously represent levels of syntactic structure and also higher-level semantic representations. 

\paragraph{Relative clauses}
We also investigate how the model behaves on relative clauses, expecting that a clause should form a constituent. We investigate sentences of the form "N that Rel VP", and count how often the relative clause is merged as a single constituent. We sample 600 sentences for the intransitive main verbs and 1,282 sentences for transitive main verbs.

\begin{table}[h]
    \def\arraystretch{1.5}
    \centering
    \begin{tabular}{|c|c c|}
    \hline
    Sentence type & [Rel.] & No [Rel.] \\
    \hline
         Sing. subject, intrans. & 4 & 596  \\
         Plur. subject, intrans. & 51 & 549 \\
         Sing. subject, trans. & 28 & 1412  \\
         Plur. subject, trans. & 125  & 1315  \\
    \hline
    \end{tabular}
    \caption{Counts for sentence parses that merge relative clauses as a constituent and parses that split relative clauses between constituents}
    \label{tab:relative}
\end{table} 

We observe the model very rarely creates a constituent for a relative clause, instead splitting the elements of the relative clause across constituents. Again, this is highly surprising from a linguistic standpoint, where the recursive substitutibility of relative clauses is used as strong evidence that they form a single constituent.

\begin{figure}[!b]
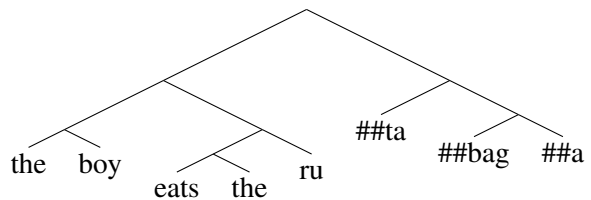

    \centering
    % \includegraphics{Tree.eps}
    % \Tree [.S [.NP \qroof{the dogs}.NP ][.VP run(s)]]
    \Tree [. [. [. the boy ][. [. eats the ] ru ] ] [. \#\#ta [. \#\#bag \#\#a ] ] ]
    \caption{Tree with split subword tokens}
    \label{fig:subword_tree}
\end{figure}

\begin{figure*}[!htb]
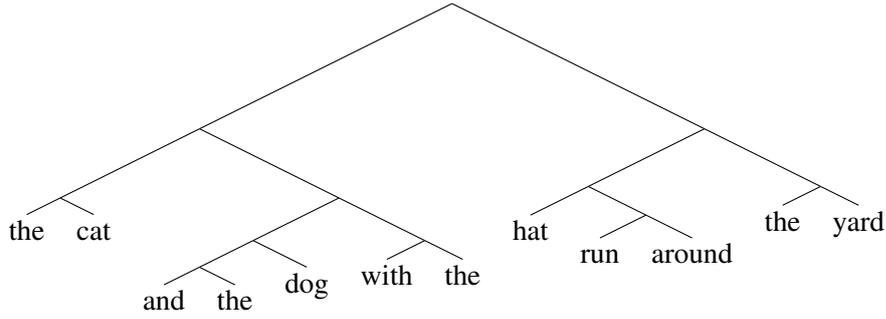

    \centering
    % \includegraphics{Tree.eps}
    % \Tree [.S [.NP \qroof{the dogs}.NP ][.VP run(s)]]
    \Tree [.  [.  [ the cat ] [ [ [and the ] dog ] [ with the ]] ]   [ [ hat [ run around ] ] [ the yard ] ] ]
    \caption{Example tree from the Tree Transformer model}
    \label{fig:example}
\end{figure*}

\paragraph{Subwords}

As data is tokenized using WordPiece, long words may be split into subword tokens–for instance, the word \textit{rutabaga} is represented as $[ru, \#\#ta, \#\#bag, \#\#a]$. We would expect these tokens be immediately recomposed by Tree Transformer, however, we often see trees such as \autoref{fig:subword_tree}, where tokens may be split across constituents.

\paragraph{Analyzing an example tree}
We provide an example tree for a complex sentence in \autoref{fig:example}.

The tree demonstrates how the model's representations often diverge significantly from linguistic insights; for example, splitting "with the" and "hat" into completely distinct constituents seems highly unexpected. However, other constituents, such as "the cat", "run around" and "the yard", align more closely with traditional theories. As these constituent representations appear to have low confidence, high variability, and inexplicable tendencies, there is little evidence to believe that the model is learning meaningful structures.  

\section{Predicting Agreement Violations}
\label{sec:agreement}
In the second set of experiments, we explore the performance of Tree Transformer models and standard Transformer models on error detection tasks that require constituent representations.

We finetune models to predict whether a sentence contains an agreement violation, focusing on number agreement between a noun phrase and its corresponding verb. While this task is trivial on simple sentences ("he/they runs/run"), it becomes more difficult for sentences with \textit{embedded clauses} and \textit{long-distance dependencies}, such as "the man who walks the dogs runs", and syntactic models that rely solely on close-proximity effects would struggle to make good predictions.

We utilize three different settings of the task that are increasingly ambiguous and require more robust generalizations, forming more and more deeply embedded sentences in order to mislead models that make incorrect generalizations.

\subsection{Data}
\label{sec:generating-violations}
We generate English sentences using a probabilistic context-free grammar (PCFG) and a small set of English words. The PCFG also optionally adds relative clauses in a recursive manner, possibly having deep embedded clauses. The full PCFG is specified in \autoref{sec:pcfg}. For example, our PCFG generates the valid sentences in \autoref{tab:valid}.

\begin{table}[h]
    \def\arraystretch{1.5}
    \centering
    \begin{tabularx}{\columnwidth}{|X|}
    \hline
         we kiss a duck  \\
         the rabbit that walks punches me \\
         the shiny cat punches these rutabagas that kick us and thinks \\
    \hline
    \end{tabularx}
    \caption{Valid sentences generated by our PCFG}
    \label{tab:valid}
\end{table} 

In order to generate sentences with agreement violations, we randomly select a verb from a valid sentence and switch the number, making singular verbs plural and vice versa. This gives incorrect sentences such as those in \autoref{tab:invalid}

\begin{table}[h]
    \def\arraystretch{1.5}
    \centering
    \begin{tabularx}{\columnwidth}{|X|}
    \hline
         we kisses a duck  \\
         the rabbit that walks punch me \\
         the shiny cat punches these rutabagas that kicks us and thinks \\
    \hline
    \end{tabularx}
    \caption{Invalid sentences created by switching the number of valid sentences}
    \label{tab:invalid}
\end{table}

We train models with 2.4k training sentences and 800 sentences each for evaluation and testing. All data splits are balanced to have an even number of positive and negative examples.

\subsection{Experimental Settings}
We test models under three experimental settings, in order to analyze hierarchical and recursive generalization. We experiment with in-distribution performance, hierarchical generalization, and recursive hierarchical generalization.

\paragraph{In-distribution (\textsc{ID})}
In the in-distribution setting, we generate valid and invalid sentences using the process described in \autoref{sec:generating-violations}. We generate data identically for the training data and evaluation and testing sets. Data is available on HuggingFace\footnote{\url{https://huggingface.co/datasets/michaelginn/latent-trees-agreement-ID}}.

This setting requires in-distribution learning, where models must learn to predict agreement violations after seeing similar sentences. These sentences have a low level of recursion. We define the \textbf{average depth} of sentences by counting the maximum number of recursively embedded clauses, where a depth of 0 means the sentence has no embedded clauses. In the in-distribution setting, sentences have an average depth of 0.2. Though the data is generated by a context-free grammar, it is effectively bounded to a finite depth, so \citet{yao_self-attention_2021} would predict that self-attention models should learn the pattern well.

\paragraph{Hierarchical Generalization (\textsc{Gen})}
In the hierarchical generalization setting\footnote{\url{https://huggingface.co/datasets/michaelginn/latent-trees-agreement-GEN}}, we follow the approach of \citet{mccoy_does_2020, petty_Transformers_2021, wilson-frank-2023-inductive} and use test sets that are not identically distributed to the training and eval sets. For the training and eval data, we filter the valid sentences produced by the PCFG to only sentences which are consistent with both a linear and hierarchical heuristic, as described in \autoref{sec:heuristics}. Similarly, we filter invalid sentences to only those which are invalid under either heuristic. Some examples are shown in \autoref{tab:gen-train}. We use the same size splits as the \textsc{ID} setting.

\begin{table}[h]
    \def\arraystretch{1.5}
    \centering
    \begin{tabularx}{\columnwidth}{|X | c|}
    \hline
    Sentence & Valid? \\
    \hline
         he walks and kisses a physicist & Y  \\
        the linguist that loves a bird runs & Y \\
         the cats that thinks runs & N \\
         the dogs that hugs the ducks barks & N \\
    \hline
    \end{tabularx}
    \caption{Training sentences consistent with either heuristic}
    \label{tab:gen-train}
\end{table}

In the testing set, we explicitly filter only examples that are explained by a hierarchical heuristic. That is, we filter to valid sentences that are invalid under the linear heuristic, and invalid sentences that are valid under the linear heuristic. Some of these sentences are given in \autoref{tab:gen-test}

\begin{table}[h]
    \def\arraystretch{1.5}
    \centering
    \begin{tabularx}{\columnwidth}{|X | c|}
    \hline
    Sentence & Valid? \\
    \hline
         the phycisist that kisses the ducks laughs & Y  \\
        the turtle that loves these happy cats ponders & Y \\
         a dude that fights these rabbits run & N \\
         the cats that love the rabbit punches us & N \\
    \hline
    \end{tabularx}
    \caption{Testing sentences only explained by hierarchical heuristic}
    \label{tab:gen-test}
\end{table}

This setting provides evidence consistent with either heuristic, testing whether models tend to select a linear or hierarchical heuristic. These testing sentences tend to be deeper (as relative clauses can cause issues for the linear heuristic), with an average depth of 0.9.

\paragraph{Recursive Generalization (\textsc{Rec-Gen})}
Finally, in order to properly evaluate the generalization ability of pretrained models, we test on sentences unlike those that appear in the pretraining corpora\footnote{\url{https://huggingface.co/datasets/michaelginn/latent-trees-agreement-GENX}}. While pretrained models have likely been trained on long sentences with embeddings, the test set for this setting contains sentences with embedded clauses many times, beyond what is likely to occur in any organic corpus.

We use the same training and evaluation sets as the previous setting. For the testing set, we modify the PCFG weights to heavily produce embedded clauses, resulting in a dataset with average depth of 8.5. These sentences tend to be unrealisticly long and deep, but nonetheless follow the same agreement rule described earlier. Some examples are given in \autoref{tab:rec-test}.

\begin{table}[h]
    \def\arraystretch{1.5}
    \centering
    \begin{tabularx}{\columnwidth}{|X | c|}
    \hline
    Sentence & Valid? \\
    \hline
         those dudes that kick a dude that punches those dudes that love a cat that kicks the cheese that laughs hug a small linguist that hugs those cats that hug us & Y  \\
         a turtle that kisses these cats that punch the physicist that punches a bird that kisses us and think love you & N \\
    \hline
    \end{tabularx}
    \caption{Testing sentences requiring hierarchical heuristic, and deeper than real sentences}
    \label{tab:rec-test}
\end{table}

While a human may not be able to validate these sentences accurately upon hearing them, we believe they would be able to label sentences correctly given written text (and a pencil and paper), as they understand how to apply the rule recursively. This setting allows us to evaluate pretrained models on data truly unlike their training data, and thus to fairly evaluate generalization ability.

\subsection{Models}
We compare four model variations, based on the BERT architecture \citep{devlin2019bert}. Specifically, we are interested in evaluating the effects of hierarchical attention mechanisms and pretraining. For all models, we use the HuggingFace transformers implementation\footnote{\url{https://huggingface.co/docs/transformers/model_doc/bert}}. 

\paragraph{BERT}
We use a standard BERT model (\textsc{bert-base-uncased}), randomly initialized, as a baseline. We train the model on the corresponding training set for each experimental setting, without any pretraining. As \citet{mueller_coloring_2022, mueller_how_2023} suggest that pretraining imparts hierarchical bias, we also compare with the standard pretrained BERT English model, fine-tuned on the training data.

\paragraph{TreeBERT}
We utilize BERT-style Tree Transformer models (TreeBERT), as in \autoref{sec:pretraining}. We use a randomly initialized TreeBERT model and the pretrained language model from \autoref{sec:pretraining}. 

\subsection{Training}
\begin{table}[htb]
    \def\arraystretch{1.5}
    \centering
    \begin{tabularx}{\columnwidth}{|X | c|}
    \hline
    Parameter & Value \\
    \hline
    Optimizer & AdamW  \\
    $\beta_1$ & 0.9 \\
    $\beta_2$ & 0.999 \\
    $\epsilon$ & $1\text{E}{-8}$ \\
    Learning rate & $2\text{E}{-5}$ \\
    Batch size & 192 \\
    Max Epochs & 100 \\
    GPU & NVIDIA V100 \\
    Early stopping patience & 3 \\
    Early stopping delay & 50 epochs \\
    \hline
    \end{tabularx}
    \caption{Training hyperparameters}
    \label{tab:hyperparameters}
\end{table}
We train models using the hyperparameters given in \autoref{tab:hyperparameters}, and early stopping based on the evaluation set. Training generally took around 20 minutes per model.

\subsection{Results}
Our results across three different settings and three models are given in \autoref{tab:results}. We run models for ten trials each with different random seeds.

\begin{table*}[htb]
    \def\arraystretch{1.5}
    \centering
    \begin{tabular}{| c | c c c | c c c | c c c |}
    \hline
     & \multicolumn{3}{c |}{\textsc{ID}} & \multicolumn{3}{c |}{\textsc{Gen}} & \multicolumn{3}{c |}{\textsc{Rec-Gen}} \\
    Model & P & R & F1 & P & R & F1 & P & R & F1 \\
    \hline
    BERT & 95.9 & 95.9 & 95.9 & \textbf{68.5} & \textbf{66.4} & \textbf{67.4} & 57.1 & 55.3 & 56.1 \\
    TreeBERT & \textbf{96.6} & \textbf{96.5} & \textbf{96.6} & 67.1 & 65.0 & 66.1 & 57.2 & 54.7 & 55.9 \\
    BERT (pretrained) & 96.1 & 96.0 & 96.0 & 66.4 & 64.6 & 65.5 & 55.5 & 53.7 & 54.6 \\
    TreeBERT (pretrained) & 96.2 & 96.1 & 96.2 & 68.0 & 65.6 & 66.7 & \textbf{58.7} & \textbf{56.1}  & \textbf{57.4} \\
    \hline
    \end{tabular}
    \caption{Average precision, recall, and F1 score     for different models across four experimental settings after 10 trials}
    \label{tab:results}
\end{table*}

In the in-distribution setting, the Tree Transformer model slightly outperformed the randomly initialized and pretrained BERT models (+0.7\%, $p<0.0001$). In the hierarchical generalization setting, the randomly initialized BERT model performed the best, while the Tree Transformer underperformed (-1.3\%). In the recursive generalization setting, the pretrained Tree Transformer model outperforms the BERT models (+1.3\%, $p<0.0001$). 

\subsection{Discussion}
\subsubsection{Comparison of Models}
\paragraph{In-Distribution Setting}
As expected, all models perform very well on the task in the in-distribution setting. This supports the findings of \citet{yao_self-attention_2021} that Transformer models can learn context-free grammars as long as they are bounded in depth. 

The Tree Transformer models outperforms the baseline, but the change is small (although statistically significant). In general, it seems likely that Tree Transformers and standard transformers struggle on the same types of data points (the incorrect 3.4\%). We also see little effect of pretraining across either architecture, which fails to support the findings on \citet{mccoy_does_2020}. In particular, we would expect that language model pretraining would provide a significant benefit to Tree Transformers, helping induce highly meaningful constituent structures. 

\paragraph{Hierarchical Generalization Setting}
In the hierarchical generalization setting, models must choose a hierarchical rule over a linear one given ambiguous training data. Prior work \citep{mccoy_does_2020, petty_Transformers_2021, wilson-frank-2023-inductive} has indicated that this is difficult for standard transformers. Our findings support this result, with the standard BERT model showing a significant drop in F1 score from the previous setting.

Surprisngly, the pretrained BERT model slightly underperforms the randomly initialized BERT model, again contradicting research \citep{mueller_coloring_2022, mueller_how_2023} which argues that pretraining helps impart a hierarchical bias. 

\citet{mccoy_does_2020} finds that models given explicit hierarchical inputs can make hierarchical generalizations more readily. Here, we find that the Tree Transformer model, which uses an implicit hierarchical representation, does not necessarily make these generalizations as well, underperforming the baseline. If we suppose that models with explicit parse trees should perform well on this task, then this result indicates that the Tree Transformer is not necessarily inducing a useful parse tree, supporting our findings in \autoref{sec:pretraining}. Overall, none of the models we tested were able to readily make hierarchical generalizations, and future work should continue to examine additional approaches.

\paragraph{Recursive Generalization Setting}
In the recursive generalization setting, models must make hierarchical generalizations on sentences much deeper than any pretraining or training data. This helps avoid the possibility that the pretrained model has seen sentences of the kind in our test set during pretraining. 

All models again demonstrate significant drops from the prior setting, indicating that while these models may learn hierarchical generalizations, they are unable to learn recursive generalizations and make predictions on deeper sentences than they were trained on. This result suggests that the effect of pretraining, as noted in \citet{mueller_coloring_2022, mueller_how_2023}, may be simply because models were pretrained on data similar to the test sentences in a generalization experiment.

The pretrained Tree Transformer model outperforms the randomly initialized models by a small margin and the pretrained BERT model by 2.8\%. On one hand, our pretrained Tree Transformer was trained on an order of magnitude less data than BERT, and full pretraining may demonstrate greater improvements. On the other hand, the benefits we see may simply be the result of the well-studied effect that inducing \textit{any} structure–even random–to unstructured data can aid in machine learning tasks \citep{rahimi2007random, breiman2001random}.

\subsubsection{Tree Transformer}
In general, the Tree Transformer model did not fully overcome the challenge of generalizing a hierarchical and recursive rule beyond training data, only outperforming by small margins in two of the three settings. Furthermore, since we pretrained a Tree Transformer model on a large enough corpus to induce high-quality language modeling, it is unclear what would be necessary to train a Tree Transformer with highly effective and meaningful constituent structures.

These results seem to point to fundamental, methodological issues with Tree Transformers. While the modified attention mechanism certainly induces a method to extract hierarchical structure, these structures seem to provide little benefit on structured tasks (nor linguistic insights, as discussed in \autoref{sec:pretraining}). We suggest a few key limitations.

First, Tree Transformer learns a single constituent structure, which limits how much tokens can interact outside of their constituents at each level. We are well aware of strong long-distance dependencies occurring across many languages, and the Tree Transformer mechanism would seem to interfere with the ability to learn these relationships. 

Second, Tree Transformer induces a syntactic structure over hidden weights, where the highest level of weights should represent the entire sentence unit, branching downward. However, research has demonstrated that BERT embeddings learn syntactic \textit{and} semantic relationships, transitioning from the former to the latter in increasing layers \citep{jawahar-etal-2019-bert}, and enforcing syntactic constituency across all layers may interfere with this ability.

Third, while Tree Transformers are biased to utilize hierarchical representations, they do not necessarily have recursive ability. Unlike recursive neural networks (as in \autoref{sec:other_models}), Tree Transformers are only applied once to input strings, regardless of length. This means that in order to handle sentences of varying depth, multiple attention layers must be able to learn the same mapping (as the relevant tokens may be constituents at any depth). In the future, we wish to compare these approaches with a recursive architecture such as \citet{chowdhury_modeling_2021}, in which the model is repeatedly applied based on the sentence depth.

In general, the main effect of the Tree Transformer modification seems to be to reduce long-distance attention at the lowest layers, which seems to have minimal benefits to language understanding tasks. \citet{Wang2019TreeTI} found benefits of Tree Transformer on language modeling on a small corpus and syntactic parsing, but these improvements were generally small and could be the result of inducing random structure. 

\subsection{Limitations}
This work demonstrates the limited benefits of the Tree Transformer architecture at learning meaningful or useful representations for English data. However, we examine a single pattern, English subject-verb agreement, and perform a limited investigation of latent representations, and it is possible that Tree Transformer has greater benefits for another language or a certain type of language pattern.

Additionally, this work uses artifically generated data that does not exactly resemble organically-produced language. We believe that agreement rules are strongly followed by native speakers, and that our data is thus a good approximation of real language.  

\section{Conclusion}
We explore the Tree Transformer architecture, which use a modified attention mechanism to create a hierarchical inductive bias, pretraining a Tree Transformer model on a large English corpus. We investigate the latent tree structured representations, finding little evidence to believe that the structures represent meaningful relationships, whether similar to standard syntactic theories or not.

We evaluate models on their ability to predict agreement violations, requiring hierarchical and recursive generalizations, comparing with standard BERT models. We find that the BERT models fail to learn robust generalizations, and pretraining does not necessarily impart the required hierarchical biases, contrary to prior research. The Tree Transformer models slightly outperform in certain settings; in particular, in the recursive generalization setting, our pretrained Tree Transformer model outperforms the next best model by 1.3\% and the equivalent pretrained BERT model by 2.8\%, despite being pretrained on far less data.

However, the effects are small, and could be due solely to inducing structure (though not meaningful) at all to the unstructured data. As we would expect a strong benefit from using constituency parsed data, we conclude with limited evidence that Tree Transformer is either a meaningful model of syntax, or that it is useful at syntactic tasks. 

\subsection{Future Work}
The Tree Transformer may not be the ideal approach to induce constituency relationships. We plan to experiment with models such as Gumbel-Tree LSTMS \citep{Choi_Yoo_Lee_2018}, CRvNNs \citep{chowdhury_modeling_2021}, and push-down transformers \citep{murty2023pushdown} which have explicit recursive biases with significantly different architectures. 

\bibliography{anthology,custom}
\bibliographystyle{acl_natbib}

\appendix
\section{Probabilistic Context-Free Grammar for Generating Data}
\label{sec:pcfg}
We generate English sentences using the PCFG specified in \autoref{fig:pcfg}.

\lstset{
    basicstyle=\ttfamily,
    keywordstyle=\color{blue},
    stringstyle=\itshape,
    morekeywords={S, VP_3Sg, VP, NP_3Sg_nom, NP_common_Sg, Det_Sg, NP_nom, NP_common_Pl, Det_Pl, NP_acc, N_bar_common_Sg, N_bar_common_Pl, N_common, Rel_Sg, Rel_Pl, VI, VT, Adj},
    morestring=[b]',
    mathescape=true,
    literate={->}{$\mathbf{\rightarrow}$}{2}
             {|}{$\mathbf{|}$}{1}
             {[}{{\textcolor{lightgray}{[}}}{1}
             {]}{{\textcolor{lightgray}{]}}}{1},
    morecomment=[s][\color{lightgray}]{[}{]}
}

\begin{figure*}
\begin{lstlisting}
S             -> NP_3Sg_nom VP_3Sg [0.5] | NP_nom VP [0.5]

VP_3Sg        -> VT '+s' NP_acc [0.475] | VI '+s' [0.475] | VP_3Sg 'and' VP_3Sg [0.05]
VP            -> VT      NP_acc [0.475] | VI      [0.475] | VP     'and' VP     [0.05]

NP_3Sg_nom    -> 'he' [0.25] | 'she' [0.25] | NP_common_Sg [0.5]
NP_common_Sg  -> Det_Sg N_bar_common_Sg [1]
Det_Sg        -> 'the' [0.5] | 'a' [0.5]

NP_nom        -> 'I' [0.125] | 'you' [0.125] | 'we' [0.125] | 'they' [0.125] | NP_common_Pl [0.5]
NP_common_Pl  -> Det_Pl N_bar_common_Pl [0.8] | NP_common_Pl 'and' NP_common_Pl [0.2]
Det_Pl        -> 'the' [0.333] | 'those' [0.333] | 'these' [0.333]

NP_acc        -> 'me' [0.075] | 'you' [0.075] | 'us' [0.075] | 'them' [0.075] | NP_common_Pl [0.35] | NP_common_Sg [0.35]

N_bar_common_Sg  -> Adj N_bar_common_Sg [0.2] | N_common 'that' VP_3Sg [0.2] | N_common [0.6]
N_bar_common_Pl  -> Adj N_bar_common_Pl [0.2] | N_common '+s' 'that' VP [0.15] | N_common '+s' [0.65]

N_common      -> 'girl' [0.0625] | 'boy' [0.0625] | 'cat' [0.0625] | 'turtle' [0.0625] | 'rutabaga' [0.0625] | 'duck' [0.0625] | 'cheese' [0.0625] | 'dude' [0.0625] | 'rabbit' [0.0625] | 'wug' [0.0625] | 'linguist' [0.0625] | 'physicist' [0.0625] | 'lady' [0.0625] | 'dog' [0.0625] | 'cat' [0.0625] | 'bird' [0.0625]

Rel_Sg         -> 'that' VP_3Sg [1]
Rel_Pl         -> 'that' VP [1]

VI            -> 'run' [0.2] | 'walk' [0.2] | 'think' [0.2] | 'laugh' [0.2] | 'ponder' [0.2]
VT            -> 'kick' [0.166] | 'kiss' [0.166] | 'hug' [0.166] | 'punch' [0.166] | 'fight' [0.166] | 'love' [0.166]

Adj           -> 'big' [0.125] | 'small' [0.125] | 'happy' [0.125] | 'mad' [0.125] | 'red' [0.125] | 'blue' [0.125] | 'sparkling' [0.125] | 'shiny' [0.125]
\end{lstlisting}
\caption{PCFG for generating data}
\label{fig:pcfg}
\end{figure*}

\end{document}